# Lightweight MRI-Based Automated Segmentation of Pancreatic Cancer with Auto3DSeg

Keshav Jha[1][0009-0003-2163-1290], William Sharp[1] [0009-0003-8154-9942], Dominic LaBella[2] [0000-0003-1713-9538]

[1] Duke University, Durham NC 277108, USA

[2] Department of Radiation Oncology, Duke University Medical Center, Durham NC, 27710, USA

**Abstract**

Accurate delineation of pancreatic tumors is critical for diagnosis, treatment planning, and outcome assessment, yet automated segmentation remains challenging due to anatomical variability and limited dataset availability. In this study, SegResNet models, as part of the Auto3DSeg architecture, were trained and evaluated on two MRI-based pancreatic tumor segmentation tasks as part of the 2025 PANTHER Challenge. Algorithm methodology included 5-fold cross validation with STAPLE ensembling after focusing on an anatomically relevant region-of-interest. The Pancreatic Tumor Segmentation on Diagnostic MRI task 1 training set included 91 T1-weighted arterial contrast-enhanced MRI with expert annotated pancreas and tumor labels. The Pancreatic Tumor Segmentation on MR-Linac task 2 training set used 50 T2-weighted MR-Linac cases with expert annotated pancreas and tumor labels. Algorithm-automated segmentation performance of pancreatic tumor was assessed using Dice Similarity Coefficient (DSC), 5 mm DSC, 95[th] percentile Hausdorff Distance (HD95), Mean Average Surface Distance (MASD), and Root Mean Square Error (RMSE). For Task 1, the algorithm achieved a DSC of 0.56, 5 mm DSC of 0.73, HD95 of 41.1 mm, MASD of 26.0 mm, and RMSE of 5164 mm³. For Task 2, performance decreased, with a DSC of 0.33, 5 mm DSC of 0.50, HD95 of 20.1 mm, MASD of 7.2 mm, and RMSE of 17,203 mm³. These findings illustrate the challenges of MRI-based pancreatic tumor segmentation with small datasets, highlighting variability introduced by different MRI sequences. Despite modest performance, the results demonstrate potential for automated delineation and emphasize the need for larger, standardized MRI datasets to improve model robustness and clinical utility.

**Keywords:** Artificial Intelligence, Pancreatic Ductal Adenocarcinoma (PDAC), Automated Segmentation, Auto3DSeg, SegResNet, MR-Linac, PANTHER

## 1.    Introduction

The 2025 **Pan**creatic Tumor Segmentation in **Ther**apeutic and Diagnostic MRI (PANTHER) challenge seeks to advance the use of Artificial Intelligence (AI)-driven solutions for the segmentation of both diagnostic and real time treatment Magnetic Resonance Imaging (MRI) scans in patients with pancreatic cancer. In 2025, pancreatic cancer is estimated to be responsible for 3.3% of all new cancer cases and responsible for 8.4% of all cancer deaths, with a 13.3% 5-Year Relative Survival [1]. Moreover, 50% of pancreatic cancers present at distant sites, emphasizing the importance of early diagnosis [1].

The role of imaging throughout diagnosis, intervention planning, and post-treatment evaluation is critical. Gross tumor volume (GTV) delineation is essential to radiotherapy (RT) treatment planning and clinical target volume (CTV) and prescription target volume (PTV) delineation. Due to the high anatomical variability of the pancreas, assessment of images can prove difficult [2]. A contrast-enhanced computed tomography (CT) scan is currently the primary diagnostic and GTV delineation tool, but MRI imaging offers several notable advantages including the ability to characterize liver lesions and metastases that are indeterminable or not visible on CT [3]. MRI is often utilized for individuals suspected to have an isoattenuated mass or a contraindication for contrast-enhanced CT [4]. Additionally, the inclusion of MRI in GTV delineation has been shown to reduce interobserver variability and volume, while also outlining the positions of bowel structures typically near to the tumor site [3]. Established protocols for MRI based treatment planning include MRI (1.5–3.0 T, slice thickness ≤ 3 mm) performed the same day as the planning CT with similar positioning, organ-filling, and respiration. T1-weighted images with and without contrast-enhancement should be used for GTV delineation [5].

Previous work with daily MRI-guided Online Adaptive Radiation Therapy automated segmentation of GTV for pancreatic cancer using dynamic contrast enhanced MR has shown to be comparable to expert radiation oncology GTV contours [6]. Since MRI-guided adaptive radiotherapy requires accurate and timely GTV delineation, this project aims to develop lightweight SegResNet models with STAPLE ensembling using the Auto3DSeg architecture [7,8].

## 2.    Methods

We used Auto3DSeg architecture together with the MONAI framework to create auto-segmentation models for pancreatic tumor delineation on MRI [7,8]. For Task 1, the model was trained to delineate pancreatic tumors using T1-weighted arterial contrast-enhanced MRI scans. For Task 2 the model utilized T2-weighted MRI for MR-Linac adaptive radiotherapy scans to segment pancreatic tumors.

### 2.1.    Imaging Data

The dataset used was released through the PANTHER challenge at MICCAI 2025, as available on Zenodo [9]. The dataset for Task 1 consists of 92 T1-weighted contrast-enhanced arterial phase MRI for patients with pathologically confirmed Pancreatic Ductal Adenocarcinoma (PDAC) who underwent diagnostic MRI as part of their clinical care at the Radboud University Medical Center (RUMC) in the Netherlands. Each case includes expert annotations for the pancreas and the pancreatic tumor as shown in **Figure 1**. The dataset for Task 2 contains 50 annotated T2-weighted MR-Linac images from patients acquired during radiotherapy throughout various stages of treatment at Odense University Hospital (OUH) in Denmark with expert annotations for the pancreas and pancreatic tumor as shown in **Figure 2**. Additionally, there are 367 unannotated cases in various phases, with and without contrast, and including diffusion weighted imaging series; however, we did not utilize these additional cases in our algorithm development for either Task 1 or Task 2. Imaging data and the associated segmentation labels were provided as MetaImage (.mha) files.

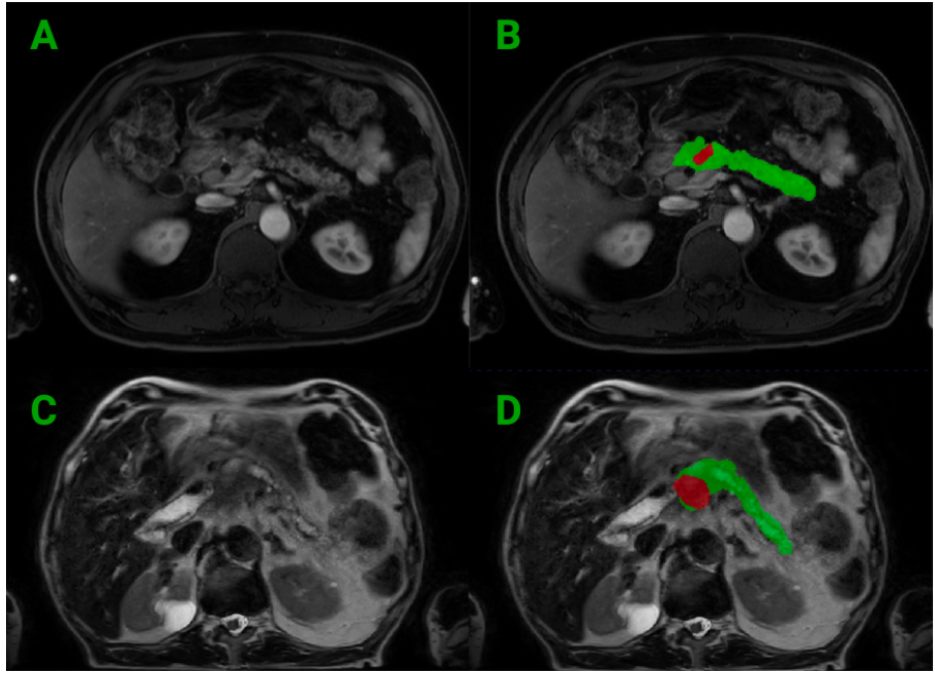

**Fig. 1.** Axial MRI T1-weighted contrast-enhanced arterial phase image (A) with expert annotated pancreas (Green) and tumor (Red) labels (B) as part of the Task 1 training set. T2-weighted MRI for MR-Linac adaptive radiotherapy (C) with expert annotated pancreas (Green) and tumor (Red) labels (D) as part of the Task 2 training set.

### 2.2. Image Pre-Processing

To improve computational efficiency, a spatial cropping strategy informed by the distribution of reference standard labels was implemented. Specifically, all training cases were analyzed to determine the minimum and maximum spatial extents of the pancreas and tumor annotations. These bounds were extended by 30 mm to define a 3D cropping region to account for inter-patient anatomical variability.

This cropping window approximates the typical spatial location of the pancreas and surrounding pathology. For Task 1, a fixed percent window crop was applied along each axis to capture the image in the 10-90% range in all axis dimensions, as seen in **Figure 2**. For Task 2, a dataset-informed percent window crop (X: 0.3280.790, Y: 0.323-0.705, Z: 0.148-1.000) was applied, as seen in **Figure 3**. This preprocessing reduces the input volume for Task 1 by 49% (keeping 51% of the voxels) and for Task 2 by 85%. By reducing the disparity between background and tumor label, we achieve more focused and efficient analysis.

### 2.3. Model Architecture and Implementation

For both Task 1 and Task 2, we utilized the Auto3DSeg framework as part of the MONAI distribution [7,8]. Utilizing Auto3DSeg allows us to leverage advanced techniques including automated hyperparameter tuning, efficient hardware utilization, and five-fold training and validation strategies. Within Auto3DSeg, we used the SegResNet network architecture, which has been shown to have improved performance over the alternative Swin UNETR and DiNTS networks as previously described by Myronenko [10,11,12]. SegResNet is a semantic segmentation model based on an encoder-decoder structure, where the number of initial feature maps is set to 32 [7,8]. The encoder is composed of five residual blocks with instance normalization. Down-sampling between stages is performed via stride-2 convolutions across five stages, consisting of 1, 2, 2, 4, and 4 convolutional layers, respectively [7,8]. We utilized the AdamW optimizer with a learning rate of 0.0002 that had a weight decay rate of 0.00001. We utilized a combined Dice + Cross-Entropy loss function over all foreground classes as implemented within MONAI [7]. Data augmentation included random flipping, random rotations, random intensity scaling and shifting per the Auto3DSeg framework defaults for MRI training images [8].

An initial phase 1 training and prediction was performed on the initial fixed-window cropped image, upon which a predicted pancreas-centered bounding box was then expanded by 30 mm in all dimensions, as seen in **Figure 2**. The purpose of this additional expansion and cropping was to allow for a more focused region-of-interest (ROI) for training and inference, to account for inter-patient anatomical variability, learn nearby anatomical features, and account for imperfect phase 1 prediction labels. For phase 1 training for Task 1 we used a training ROI of [256, 208, 56] and for Task 2 an ROI of [192, 160, 104]. For phase 2 training for Task 1 we used a training ROI of [144, 96, 40] and for Task 2 an ROI of [176, 112, 72].

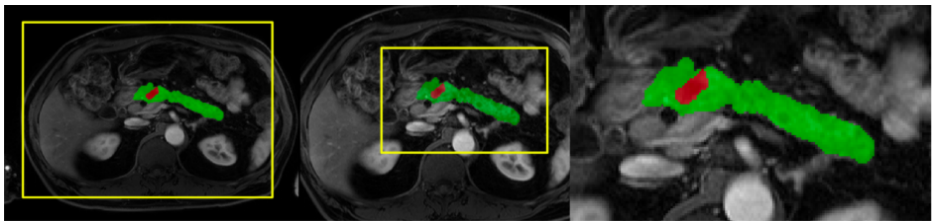

**Fig. 2.** Processing pipeline for Task 1 as shown on axial T1-weighted contrast-arterial enhanced MRI. (Left) Preprocessing fixed window crops of 10-90% (yellow box) in all dimensions for phase 1 training and inference. (Middle) Tighter crop boundaries (yellow box) determined by phase 1 inference. (Right) Final model input for phase 2 predictions.

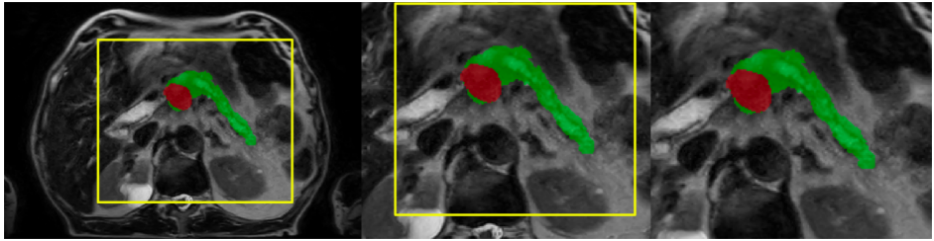

**Fig. 3**. Processing pipeline for Task 2 as shown on T2-weighted MRI for MR-Linac adaptive radiotherapy. (Left) Preprocessing fixed window crops in all dimensions for phase 1 training and inference. (Middle) Tighter crop boundaries (yellow box) determined by phase 1 inference (green pancreas). (Right) Final model input for phase 2 predictions.

A single GPU RTX 2070 Super with 8 GB of available VRAM was used to perform 5-fold training and cross-validation. However, only 4 GB of VRAM was utilized during training. Because of this lighter-weight setup, we were able to utilize the Auto3DSeg framework, but performance limitations are expected when compared to previous studies [8,10]. Due to these memory and processing speed limitations, we attempted image resizing after completion of a rough pancreas prediction on the larger image as previously mentioned in section 2.3.

## 2.4. Training Stages

No pretraining was used for these models. Training for both tasks consisted of 1000 epochs in phase 1 and 750 epochs in phase 2. To mitigate overfitting, we employed a 5-fold cross-validation system with our data randomly distributed between folds. Each fold was iteratively used as the validation set while the remaining folds were used for training. Due to limited computational power, the model was limited in terms of its batch size and ROI.

**Table 1.** Task 1 fold-wise overall accuracy for the SegResNet models.

| Class ID | Structure | Fold 1 | Fold 2 | Fold 3 | Fold 4 | Fold 5 | Mean |
|---|---|---|---|---|---|---|---|
| 0 | Background | -- | -- | -- | -- | -- | -- |
| 1 | Tumor | 0.559 | 0.588 | 0.602 | 0.644 | 0.621 | 0.603 |
| 2 | Pancreas | 0.796 | 0.756 | 0.798 | 0.797 | 0.847 | 0.799 |

**Table 2.** Task 2 fold-wise overall accuracy for the SegResNet models.

| Class ID | Structure | Fold 1 | Fold 2 | Fold 3 | Fold 4 | Fold 5 | Mean |
|---|---|---|---|---|---|---|---|
| 0 | Background | -- | -- | -- | -- | -- | -- |
| 1 | Tumor | 0.395 | 0.485 | 0.547 | 0.477 | 0.361 | 0.453 |

| 2 | Pancreas | 0.536 | 0.527 | 0.599 | 0.533 | 0.533 | 0.546 |

Following segmentation within the cropped region, the resulting predictions were mapped onto the original image space by inversing the cropping transformation done in preprocessing. This ensured that final segmentation masks aligned with the patient's full imaging volume. By restoring predictions to their native spatial coordinates, consistency with the original MRI scans was maintained.

Model selection was based on the best-performing checkpoint on the validation set rather than the final epoch model. For the final prediction, we employed the Simultaneous Truth and Performance Level Estimation (STAPLE) algorithm, which fuses predictions from multiple models [13].

### 2.5. Final Testing Evaluation

The PANTHER competition organizers used the following evaluation metrics for the final testing evaluation on a hidden dataset of 30 cases for each Task 1 and Task 2.

The Dice Similarity Coefficient (DSC) was computed to quantify the spatial overlap between the predicted segmentation and reference standard for both the pancreas and tumor. This metric is widely used in medical image segmentation challenges as a measure of agreement between two binary masks. Equation 1 shows the DSC formula, where P is the predicted segmentation and G is the reference standard segmentation [14].

$$DSC(P,G) = \frac{2|P \cap G|}{|P| + |G|}$$

(1)

The 5 mm Surface Dice measures the proportion of surface points from one segmentation that lie within a 5 mm tolerance of the other segmentation surface. Equation 2 shows the 5 mm Surface Dice, where $S_p$ and $S_g$ are the surface voxels of the predicted and reference standard segmentations, respectively. d(x,Y) denotes the shortest Euclidean distance from point x to set Y, and |.| denotes cardinality.

$$SurfaceDSC_\tau(S_P, S_G) = \frac{\left|\{p \in S_P : d(p, S_G) \leq \tau\}\right| + \left|\{g \in S_G : d(g, S_G) \leq \tau\}\right|}{|S_P| + |S_G|}, \tau = 5\,mm$$

(2)

The Mean Average Surface Distance (MASD) captures the average boundary discrepancy between the prediction and reference standard. This computes the mean of the average distance between surfaces in both directions, offering a more balanced evaluation. Equation 3 shows the MASD formula, where $S_p$ and $S_g$ denote the surface voxels of the predicted and reference standard segmentations, and d(x, Y) is the shortest Euclidean distance from a point to a set.

$$MASD(S_P, S_G) = \frac{1}{|S_P| + |S_G|} \left( \sum_{p \in S_P} d(p, S_G) + \sum_{g \in S_G} d(g, S_P) \right) \quad (3)$$

The Hausdorff Distance 95% (HD95) is computed to quantify the worst-case boundary disagreement while reducing outlier influence by considering the 95th percentile of distances rather than the maximum. Equation 4 shows the HD95 formula, where $S_p$ and $S_g$ are the surfaces of the predicted and reference standard masks, and $d(x, Y)$ is the shortest Euclidean distance.

$$HD95(S_P, S_G) = max(\text{percent}_{95}(\{d(p, S_G): p \in S_P\}), \text{percent}_{95}(\{d(g, S_P): g \in S_G\})) \quad (4)$$

To evaluate differences in tumor burden estimation, the Root Mean Square Error (RMSE) is computed between predicted and reference standard tumor volumes. Equation 5 shows the RMSE formula, where $V_P$ and $V_G$ represent the predicted and reference standard tumor volumes, calculated as the product of mask voxels (P or G) and voxel volume, N is the number of voxels, and M is the total number of test cases.

$$V_p = \sum_{i=1}^{N} P_i \cdot v_{voxel}, \; V_G = \sum_{i=1}^{N} G_i \cdot v_{voxel} \; RMSE = \sqrt{\frac{1}{M} \sum_{j=1}^{M} \left( V_p^{(j)} - V_G^{(j)} \right)^2} \quad (5)$$

The code used for evaluation can be found at https://github.com/DIAGNijmegen/PANTHER_baseline/blob/main/Evaluation/evaluate_local.py.

## 3. Results

Performance was evaluated on the open development phase test set and is shown in Table 3.

Table 3. Open development phase Task 2 scores

| Task | DSC | 5 mm DSC | HD95 (mm) | MASD (mm) | RMSE (mm$^3$) |
|---|---|---|---|---|---|
| 1 | 0.558 | 0.730 | 41.118 | 26.007 | 5164.311 |
| 2 | 0.327 | 0.502 | 20.089 | 7.159 | 17202.848 |

Overall, segmentation performance was stronger on Task 1 than Task 2. Task 1 achieved higher spatial overlap and surface conformity, while Task 2 demonstrated poorer overlap but relatively improved boundary error. Tumor burden estimation was more accurate on diagnostic scans, with substantially lower RMSE than on MR-Linac images.

## 4.    Discussion

We sought to utilize the SegResNet model within the Auto3DSeg MONAI framework to delineate pancreatic tumors in MRI scans as part of the PANTHER challenge [7,8,15]. The segmentation performance observed in this study highlights both the potential and the limitations of training deep learning models on small pancreatic cancer datasets. With only 91 annotated cases in Task 1 and 50 in Task 2, the model achieved weak overlap in Task 1 (DSC = 0.56, 5 mm DSC = 0.73) and struggled with precise boundary delineation (MASD = 26 mm, HD95 = 41 mm). Task 2 performance further declined in terms of overlap (DSC = 0.33, 5 mm DSC = 0.50), though surface-based errors were somewhat reduced (MASD = 7 mm, HD95 = 20 mm). These findings are consistent with the challenges of pancreas and tumor segmentation, where small sample sizes limit the model's ability to generalize across the wide anatomical and imaging variability seen in clinical practice. The relatively high surface distance errors suggest that while the model can localize the tumor region, it lacks robustness in capturing fine boundary details. Integrating pretraining, self-supervised learning, or semi-supervised approaches could improve segmentation performance.

Future studies should also explore the inclusion of regional gross lymph nodes (GTVn), which play a critical role in treatment planning and staging. Including nodal disease would provide a more complete assessment of the tumor burden. Furthermore, expanding automation to generate CTV labels from GTV delineations could support radiotherapy planning by accounting for microscopic disease spread, in accordance with consensus guidelines [16]. This would allow auto segmentation tools to align with clinical workflows for pancreatic cancer.

Another promising avenue lies in the use of automated segmentation predictions to support treatment selection. Tumor size and volume play an essential role in determining eligibility for neoadjuvant chemotherapy in pancreatic cancer. Automated segmentation with precise boundary delineation could serve as a standardization measurement tool to assist with this management decision, especially for borderline resectable patients who may benefit most from neoadjuvant strategies [17]. Further work should evaluate the integration of tumor volumetrics derived from automated segmentation into clinical frameworks.

An additional opportunity is to explore other models within the Auto3DSeg framework, as we focused on the use of SegResNet alone for the competition's tasks [11,12]. Exploring the use of transformer-based models capable of self-attention could allow models to detect longer-range features within the scans [12,18].

One limitation of this study is the computational resources available for training and evaluation. Specifically, we were restricted in our memory and processing speed using the RTX 2070. Although we sought out methods to reduce input volumes, these hardware limitations likely negatively affected performance optimization.

A second limitation relates to the input data typically used for tumor delineation. In clinical practice, GTV annotations are created using all available clinical information, including MRI, CT, PET scans, endoscopy reports, and expert radiologist guidance, while this study's algorithm was developed using MRI data only [3,19]. This restriction limits the model's ability to capture features from other information sources. MRI-only model performance likely underestimates what could be achieved using multi-modal imaging data.

## 5. Conclusion

In this study, we demonstrated the possibility of lightweight automated segmentation of pancreatic tumors on MRI using the Auto3DSeg framework with SegResNet. Despite a limited dataset and computational power, our models achieved modest performance, highlighting both the challenges and potential of AI-assisted delineation in pancreatic cancer. These results underscore the value of MRI-based segmentation for MR-guided radiotherapy planning and suggest that further improvements may be made through multi-modal imaging, inclusion of nodal structures, and exploration of transformer-based architectures. Overall, this work provides a foundation for automated segmentation, with potential for integration into clinical workflows where it can support more precision and efficient treatment planning.

**Data Availability.** Our code for all steps of processing and training is available, as open source, at GitHub: https://github.com/Luufu/Two-Phase-PANTHER-Challange-Auto3DSeg-Model/tree/main. All data is available as part of the PANTHER challenge [9,15].

**Acknowledgments.** We appreciate the time and effort of the PANTHER challenge organizers as well as their provision of the data for this project.

**Disclosure of Interests.** The authors have no competing interests to declare that are relevant to the content of this article.